# Can LLMs Grade Short-Answer Reading Comprehension Questions? An Empirical Study with a Novel Dataset

**Owen Henkel**[1], **Bill Roberts**[2], **Libby Hills**[3], **Joshua McGrane**[4]
Submitted: October 2023

## Abstract

Formative assessment plays a critical role in improving learning outcomes by providing feedback on student mastery. Open-ended questions, which require students to produce multi-word, nontrivial responses, are a popular tool for formative assessment as they provide more specific insights into what students' do and don't know. However, grading open-ended questions can be time-consuming and susceptible to errors, leading teachers to resort to simpler question formats or conduct fewer formative assessments. While there has been a longstanding interest in automating the grading of short-answer questions, previous modeling approaches have been technically complex, limiting their use in formative assessment contexts. The newest generation of Large Language Models (LLMs) potentially makes grading short answer formative questions more feasible, as the models are more flexible and easier to use. This paper addresses the lack of empirical research on the role of LLMs in assessment, specifically in the grading of short answer questions for formative assessments, in two ways. First, it introduces a novel dataset of short answer reading comprehension questions, drawn from a battery of reading assessments conducted with over 150 students in Ghana. This dataset allows for the evaluation of LLMs in a new context, as they are predominantly designed and trained on data from high-income North American countries. Second, the paper empirically evaluates how well various configurations of generative LLMs can grade student short answer responses compared to expert human raters. The findings show that GPT-4, with minimal prompt engineering, performed extremely well on evaluating the novel dataset (QWK 0.92, F1 0.89), reaching near-parity with expert human raters. To our knowledge this work is the first to empirically evaluate the performance of generative LLMs on short answer reading comprehension questions using real student data, with low technical hurdles to attaining this performance. These findings suggest that generative LLMs could be used to grade formative literacy assessment tasks, potentially benefiting real-world educational settings.

[1] University of Oxford, Department of Education | owen.henkel@education.ox.ac.uk
[2] Legible Labs | bill@legiblelabs.com
[3] Jacobs Foundation | libby.hills@jacobsfoundation.org
[4] University of Melbourne, Graduate School of Education | joshua.mcgrane@unimelb.edu.au

# 1. Introduction

The potential applications of Large Language Models (LLMs) in education are myriad, and grading student work in low-stakes situations such as formative assessment, is a particularly promising area. Formative assessments, can help students monitor their progress and provide teachers with a deeper understanding of student knowledge and reasoning (Black & Wiliam, 2009; Shute, 2008). Closed-response question formats such as multiple-choice and true/false, while efficient to grade, can be time-consuming to create and have other limitations, including being more indicative of test-taking strategies rather than actual comprehension, and in some cases lack of face validity (Bellinger & DiPerna, 2011; Landauer et al., 2009). In contrast, open-ended formats such as short answer questions require students to answer in their own words, and can evaluate more integrative skills such as communication, reasoning, and analysis (Morjaria et al., 2024; Shapiro et al., 2014). However students' responses to short answer can be challenging score consistently and are extremely time-consuming for teachers to mark, which limits their widespread usage, particularly in formative assessment (Magliano & Graesser, 2012).

Automatic short answer grading (ASAG) is a longstanding research area, whicg aims to develop computational methods for evaluating student responses to open-ended questions. Early ASAG models employed statistical or domain-specific neural network approaches but struggled to generalize across different domains and contexts, requiring painstaking construction for each specific task (Burrows et al., 2015). In recent years, transfer learning, a technique involving adapting a model trained on one NLP task for use on another related task has gained traction and demonstrated dramatic performance across various NLP tasks including ASAG (Brown et al., 2020; Haller et al., 2022). By leveraging the transfer learning paradigm, where a pre-trained model is fine-tuned on a specific task using a labeled dataset, researchers have been able to achieve high levels of accuracy across a wide range of ASAG tasks (Fernandez et al., 2023; Sung et al., 2019). For instance, Fernandez et al. (2023) achieved an agreement score with expert raters of 0.84 (Cohen's Kappa) on evaluating open-response reading comprehension questions, which was close to the human-to-human agreement of 0.88.

Despite this progress, there remain least two major hurdles limiting their using ASAG in real-world education settings. First, fine-tuning these models still requires non-trivial technical expertise, and gathering and labeling a sufficiently large dataset (i.e., thousands of examples) is still a serious undertaking. Second, transfer learning models often struggle to generalize their performance beyond the specific tasks or datasets they were trained on, even when the task appears to be quite similar, a problem known as domain shift (Bommasani et al., 2022; Raffel et al., 2020; Sultan et al., 2022). This lack of generalizability limits the practical benefits of using transfer learning models, as they can't be easily adapted and deployed across different educational contexts, particularly in the context of formative assessment which occurs frequently and informally.

More recently, a new generation of LLMs (e.g., GPT-4, Claude, Gemini) commonly referred to as "generative LLMs" have demonstrated enhanced usability, allowing non-technical users to make requests in natural language (i.e., prompting), as well as the ability to generalize across a wide range of tasks, often with minimal exposure to examples (Kojima et al., 2022; Ouyang et al., 2022). While these generative models have been designed to generate text based on the input provided, rather than being specifically trained for classification tasks these LLMs have shown evidence of being able to perform exceptionally well on classification tasks, through the use of few-shot prompting where the model is given a few examples as part of the system prompt (Cohn et al., 2024; Kortemeyer, 2023; Ouyang et al., 2022). The ability to adapt a model to new classification tasks (i.e. grading) with minimal training data and technical overhead could provide a more accessible and efficient way to evaluate student understanding and provide targeted feedback.

However, there are also important concerns about the use of LLMs to assist in the marking of educational assessments. Firstly, there are questions whether the LLMs' judgments align with those of human experts who typically mark student responses (Caines et al., 2023; Weidinger et al., 2022). Secondly, the non-deterministic nature of these models means that they may not always produce the same output for a given input, which can potentially undermine the reliability and assessment process, making it difficult to compare student performance across different time points or cohorts. Thirdly, LLMs are trained on vast amounts of text data that may contain inherent biases and stereotypes, which could potentially influence the model's grading decisions and lead to unfair or discriminatory assessment of student work (Baker & Hawn, 2022). To date, there has been very little empirical research on the use of generative LLMs to mark short-answers questions, highlighting the need for further investigation to better understand their potential impact and limitations.

The current study addresses the lack of empirical research on the use of generative LLMs for educational assessment in several important ways. Firstly, we introduce a publicly available dataset, the [University of Oxford & Rising Academies Ghanaian Student Reading Comprehension Short Answer Dataset](#) (subsequently referred to as the Ghana Dataset). This dataset contains the results of various reading assessment tasks administered to 130 students between the ages of 9 and 18 in Ghana. The Ghana dataset contains the original test, over 1,500 student answers to reading comprehension questions (~1,000 short answers and ~500 MC), and scores for each answer generated by expert human raters. Importantly, the fact that this dataset was developed in collaboration with an education organization in Ghana helps to broaden the scope of research on AI in education to include underrepresented groups. This dataset can serve as a resource for future researchers investigating the potential of LLMs and other AI technologies to education in low and middle-income countries (LMICs), which is essential to ensure that these less-resourced education systems can better understand the potential risks and benefits of LLMs. Secondly, this paper provides the first empirical research on the ability of the most recent generation of generative LLMs to grade short answer reading comprehension questions. Our topline finding, that the newest generation of LLMs performed extremely well, exceeding previous benchmarks and approaching human-level performance, has important implications for educational assessment in general and the formative assessment of reading comprehension in particular.

## 2. Prior Work

### 2.1 Assessing Reading Comprehension

Depending on the context reading comprehension may refer to everything from the ability to understand the literal meaning of a passage to reasoning about the information presented in a text to evaluating the persuasiveness of claims made in a text. For the purposes of this paper, we adopt the definition of reading comprehension proposed in the PISA for Development Reading Framework (2018), which is summarized below in Figure 1. We choose this framework because it articulates increasingly sophisticated levels of reading comprehension in terms of the associated student capabilities and is focused on the assessment of reading comprehension in LMICs.

| Process | Readers are able to… |
|---|---|
| **Access and Retrieve Information** | • Identify in print individual words that would occur in the everyday listening lexicon of average adult speakers of the language.<br>• Combine words to parse sentences and represent their literal meaning.<br>• Locate explicitly stated individual pieces of information, in large passages. |
| **Literal Comprehension** | • Comprehend explicitly stated information that may be found in sentences or passages.<br>• Combine the meaning of small sets of sentences in order to form an internal representation of simple descriptions or narrations.<br>• Form a representation of the information contained across multiple sentences, connecting the idea units and structuring them in memory. |
| **Form a Broad Understanding** | • Consider the text as a whole or in a broad perspective.<br>• Demonstrate initial understanding by identifying the main topic or message or by identifying the general purpose or use of the text.<br>• Connect various pieces of information to make meaning, comparisons of degree, or understanding cause and affect relationships. |
| **Develop an Interpretation** | • Go beyond the literal meaning of a passage or a text and identify the underlying assumptions or implications.<br>• Identify and list supporting evidence, and compare and contrast information, draw together two or more pieces of information from the text |
| **Reflect on and Evaluate the Content of a Text** | • Connect information in a text to knowledge from outside sources.<br>• Assess the claims made in the text against their own knowledge of the world.<br>• Articulate and defend their own points of view |
| **Reflect on and Evaluate the Form of a Text** | • Consider a text objectively and evaluate its quality and appropriateness.<br>• Evaluate how successful an author is in portraying some characteristic or persuading a reader. |

**Figure 1** Adapted from PISA for Development Reading Framework (2018)

The first three sub-processes—accessing and retrieving information, literal comprehension, and forming a broad understanding—and the associated descriptions of student abilities closely correspond both with definitions of functional literacy and the types of cognitive abilities required by reading comprehension questions targeted for emerging readers (Keenan et al., 2008; Kirsch & Guthrie, 1977). For this reason, when discussing reading comprehension in subsequent sections we are referring to these three sub-processes and their associated skills.

Closed response questions, including multiple-choice, true/false, and fill-in-the-blank, have been the dominant method for assessing reading comprehension for several decades, likely due to their cost-effectiveness and simplicity to administer and score (Pearson & Hamm, 2006). However, they also have several limitations. First, they are time-consuming to develop, and second, students' responses may be more related to their background knowledge or test-taking strategies than to actual comprehension. There are examples of individuals performing significantly better than chance when answering the questions without even having read the corresponding passages (Cain & Oakhill, 2006; Magliano & Graesser, 2012; van den Bergh, 1990).

These limitations may become more serious in LMICs, where students can develop a mix of component reading skills that would be uncommon in monolingual, well-resourced educational settings (Klaas & Trudell, 2011; Spaull et al., 2020). For instance, a student who has adequate decoding skills but lacks sufficient understanding of the language they are reading, or one with adequate decoding and comprehension skills but a lack of familiarity with multiple-choice question formats, might score similarly to a student with weak decoding skills. Indeed, one of the main arguments for open-ended formats is that they can evaluate different aspects of reading comprehension than closed response questions can (Pearson & Hamm, 2006).

Open-ended comprehension questions require students to read (or listen) to a passage and then answer questions, recall key elements, or construct a summary in their own words (Cain & Oakhill, 2006; Shapiro et al., 2014). Unlike closed response questions, there is no single correct answer, and grading is usually conducted by experts or graders using a rubric (Reed & Vaughn, 2012). Some researchers claim that open-ended questions are appealing because the close correspondence with real-world applications of reading comprehension, i.e., face validity (Nation et al., 2010). Short answer or constructed response items require students to read a passage and then answer a specific question about the passage using their own words in a few sentences (Burrows et al., 2015). Correct answers are seen as prima facie evidence of comprehension, as there is no way of using the process of elimination to identify the most plausible of a set of options (van den Bergh, 1990).

## 2.2 Automatic Short Answer Grading

Automatic short answer grading (ASAG) has been an active area of research for over a decade, and it is beyond the scope of this study to cover it in detail. Burrows et al, provide a comprehensive overview of approaches up until and 2015; while Haller et. al., (2022) discuss more recent developments up to - but not including- the debut of the latest generation of LLMs.

The key rationale for ASAG, is that while instructors often prefer the type of information they can get from open-ended questions, grading student responses individually is a laborious process, potentially resulting in overburdened educators and compromised feedback quality students' comprehension and foster critical engagement with the subject matter (Matelsky et al., 2023). As a result, some instructors opt for simpler formats like multiple-choice questions, which provide instant feedback but lack personalized and insightful remarks (Burrows et al., 2015; Magliano & Graesser, 2012). Is however, short answer questions could be graded automatically it would be possible for teachers to get the rich feedback they need to best inform instruction without creating an unsustainable workload.

As noted in Haller et. al., (2022), over the last few years there has been a move from models based on handcrafted features to models using word-embedding and or representation learning approaches. However, the majority models used for ASAG regardless of the paradigm, are similar in that that they are explicitly trained or fine-tuned for particular grading tasks (Kortemeyer, 2023). Despite the technical complexity of developing ASAG models, the past few years has seen significant progress with narrowly scoped ASAG tasks. For instance, Sultan et al. (2016) made an important contribution by building a model that represents each sentence as the sum of the individual word embeddings and achieved state-of-the-art performance (at the time) on the SemEval benchmarking dataset. However, the dependence on prompt-specific training data often meant that it was necessary to re-train the model for each induvial short answer promote prompt, which was costly, time-consuming, and in most cases simply infeasible.

An important more recent advance in ASAG has been fine-tuning pre-trained LLMs, an approach often referred to as transfer learning. Transfer learning is a paradigm of machine learning that typically consists of two steps: pre-training and fine-tuning. During pre-training, a neural network model conducts unsupervised learning on a large-scale general dataset to establish model weights, then during the fine-tuning phase, the model is trained using supervised learning on a smaller, task-specific labeled dataset (Kojima et al., 2022). This process significantly reduces the amount of labeled data required for training and enables the model to adapt to the nuances and specifics of the target task and domain using only a few hundred labeled examples, as opposed to the thousands of examples that would have been required using previous approaches (Brown et al., 2020).

Early attempts at using the transfer learning paradigm for short-answer grading include Sung et al. (2019), who experimented with fine-tuning BERT, a widely used pretrained transformer-based language model, to classify pairs of student answers as correct or incorrect. They found that this approach produced superior results across multiple domains, reporting a 10% improvement over previous state-of-the-art results on the SemEval dataset. It was also able to classify student responses almost at the human-level agreement. More recently, in the specific case of evaluating open-response reading comprehension questions, we consider the state-of-the-art in this task to be a winning submission to the NAEP automated scoring challenge for reading comprehension (Fernandez et al., 2023). Using a BERT-based model, they achieved an agreement score with expert raters, as measured by Cohen's Kappa, of 0.84, where human-to-human scores were 0.88.

While using pre-trained language models fine-tuned with relatively smaller sets of task-specific data has become the preferred approach in ASAG and many other NLP tasks, the practical application of transfer learning models for ASAG, particularly for formative assessment in real-world educational settings, remains limited. This is largely due to a few central constraints of this approach: the technical complexity of the fine-tuning process, the continued (albeit small) need for task-specific data, and these models' difficulty in generalizing.

## 2.3 Challenges with Transfer-Learning Based Approaches

While transfer-learning based approaches have demonstrated high accuracy when evaluating specific tasks, and represent substantial improvement relative to preceding approaches, their development requires significant technical expertise, which is often unavailable in most schools. Even in situations where a school may have access to such expertise and can develop such a model, a labeled dataset of a few hundred examples of correct and incorrect student answers is required. Furthermore, the model must be fine-tuned each time the question students answer changes (Mishra et al., 2022; Ye et al., 2021). This is an example of a well-documented challenge of using the transfer learning paradigm in an educational context: domain shift (Camus & Filighera, 2020; Perez et al., 2021). Prior research on ASAG and the closely related area of Automatic Essay Scoring (AES) has found that while prompt-specific fine-tuning for essay scoring has achieved a high level of performance, it has been ineffective when applied in a cross-prompt setting. In other words, a system trained using only responses from one prompt may not generalize well to answers written in response to another prompt, even when the prompts are similar in nature (Alikaniotis et al., 2016; Ridley et al., 2020). Domain shift can be particularly challenging to mitigate because even tasks that seem quite similar in their structure and goals may have subtle differences that may not be immediately apparent but significantly impact the model's performance (Bommasani et al., 2022; Zhao et al., 2020).

For example, in the context of the ASAP dataset, which contains short student essays (150-500 words) in response to 8 different questions, models that achieved a high level of performance in grading student responses to one prompt - reading an excerpt from a short story and then explaining why the author chose to conclude the story in the way that she did - were not able to successfully grade student responses to a similar but distinct prompt - reading an excerpt from a short story and then describing the mood created by the author (Ridley et al., 2020)

In the context of ASAG, one example of the challenge of domain shift is the Fernandez et al. (2023) winning submission in the NAEP short answer competition. Their winning model, despite implementing many sophisticated approaches to increase generalizability, decreased from a Quadratic Weighted Kappa (QWK) of 0.84 on the hold-out data used for model evaluation to below 0.50 on a separate dataset NAEP developed to test for generalization (personal correspondence). This is a typical result for many auto grading models, so while transfer-learning models appear to be capable of effectively grading large-scale, formal assessments because they happen infrequently and involve grading thousands of similar student responses, they are unlikely to be practical for classroom-based assessment, including formative assessment, which occurs frequently and often informally. These contexts require models that can be easily adapted to various contexts, with limited need for new labeled data.

## 2.4 Potential of Generative LLMs for ASAG

The more recent generation of Large Language Models (LLMs), starting with ChatGPT and subsequently including a wide variety of models such as GPT-4, Claude, Llama, Mistral, Gemini, etc. (commonly referred to as "generative LLMs"), were initially trained in a manner like previous generations of LLMs. However, these models utilized significantly larger datasets and dramatically increased the number of parameters, in some cases by more than an order of magnitude (Chowdhery et al., 2023; Stiennon et al., 2022). Crucially, following the pre-training stage, these models underwent "instruction fine-tuning" to align their output with human preferences and enhance their ability to follow human instructions (Stiennon et al., 2022). While the specifics of this process vary slightly between models, the common result was greatly enhanced usability, allowing non-technical users to make requests in natural language (i.e., prompting), as well as a remarkable ability to generalize to new tasks, often with minimal exposure to examples (Ouyang et al., 2022).

Importantly, these models can interpret human-written natural language instructions, enabling users to adapt a model to new tasks simply by modifying the prompts they sent to the model, rather than requiring further training or fine-tuning. Generative LLMs have exhibited the ability to perform various linguistic tasks that previously required the use of task-specific, fine-tuned LLMs (Kojima et al., 2022; Wei et al., 2022). Moreover, there is growing evidence that these models can complete evaluation tasks on novel datasets with only minimal prompt engineering (Gilardi et al., 2023; Kuzman et al., 2023). Although the models are not explicitly trained for it, they have demonstrated impressive capabilities in classification tasks and there is mounting evidence that they can be used for certain types of grading tasks (Kortemeyer, 2023). In the specific case of using generative LLMs for grading, instead of fine-tuning custom head for a pre-trained LLM using a task specific dataset, a user can now simply write an explanation of how they wish the model to classify (i.e. grade) student answer, typically accompanied by a few examples

While there has been a fair amount of work on using LLMs for scoring essays, there is relatively little research on using generative LLMs to evaluate short-answer student responses (Kortemeyer, 2023; Mizumoto & Eguchi, 2023; Schneider et al., 2023). Morjaria et al. (2024) found that ChatGPT performed similarly to a single expert rater when marking short-answer assessments in an undergraduate medical program, but their study only marked 6 distinct questions with 10 student answers each. A review by Schneider et al. (2023) considered x and concluded that "while 'out-of-the-box' LLMs provide a valuable tool to offer a complementary perspective, their readiness for independent automated grading remains a work in progress." Cohn et al., (2024) explored short-answer responses to formative assessments in high school science and found that GPT-4 successfully scored student answers using a human-in-the-loop approach combining few-shot and active learning with chain-of-thought reasoning. However, Kortemeyer (2023) found that while LLMs could be useful for preliminary grading of introductory physics assignments, they fell short in certain aspects.

In all the above cases, the studies were conducted with a small sample size of student answers, typically in the range of dozens. Furthermore, the focus of these studies was primarily on high school and university students, with little exploration of the applicability of generative LLMs in evaluating short-answer responses at other educational levels, such as elementary or middle school. This limitation in the scope of research can be attributed, in part, to the limited number of publicly available short-answer datasets.

## 2.5 Overview of Existing Short Answer Datasets

There are various publicly available datasets commonly used for evaluating ASAG modes. Some of the most well-known includes SciEntsBank which consists of responses of students across North America in grades 3 to 6 to questions in standardized assessment, and Beetle which evaluates high-school students' interactions with a tutorial system on physics and electronics, containing 47 questions with 366 responses each. Other datasets include ASAP which provide a broad range of subjects, from the sciences to reading comprehension, but are tailored to the high school students in the US. The CREG and CREE datasets concentrate on the language learning data rather than literacy. Other datasets, such as CS and CSSAG, focus on niche subjects like computer science at the university level.

**Table 1**

*Summary of Publicly Available Short-Answer Datasets*

| Corpus | Prompt Type | Language | Learner Population | Scoring Labels |
| --- | --- | --- | --- | --- |
| **ASAP** | Sciences, biology, reading comprehension | English | High school students | Numeric [0, 1, 2, (3)] |
| **ASAP-DE** | Sciences | German | Crowdworkers | Numeric [0, 1, 2, (3)] |
| **Beetle** | Physics, fundamentals of electricity and electronics | English | High-school students | N/A |
| **CREE** | Reading comprehension for language learning | English | University students | Binary & diagnostic |
| **CREG** | Reading comprehension for language learning | German | US university students learning German | Binary & diagnostic |
| **CS** | Computer science questions | English | University students | Numeric [0, 0.5, ..., 5] |
| **CSSAG** | Computer science | German | University students | Numeric [0, 0.5, ..., 2] |
| **Powergrading** | Immigration exams | English | Unknown (crowdworkers) | Binary |
| **PT_ASAG** | Biology | Portuguese | 8th & 9th grade students | Numeric [0, 1, 2, (3)] |
| **SciEntsBank** | Science questions | English | 3rd - 6th grade students | Numeric [2, 3, 5] |
| **SRA** | Science questions | English | High school students | Entailment labels (binary & diagnostic) |

As can be seen in Table 1 above, the datasets remain predominantly focused on North American and European contexts, and lack data that focuses on foundational educational skills, such as basic reading comprehension. Our research attempts to fill the gap by creating a dataset containing short answer response to reading comprehension questions from students in Ghana. Additionally, the new dataset stands out by providing a detailed demographic breakdown, which is often absent in other datasets.

## 3. Current Study

11The goal of this study was twofold: (1) to create and share a novel dataset of short answer reading comprehension questions, and (2) to evaluate the performance of generative LLMs and various prompting strategies on grading this novel dataset.

The novel dataset introduced in this paper is important because it addresses the scarcity of diverse and comprehensive datasets of student short answer questions, particularly from students in low and middle-income countries. By making the dataset publicly available we hope to facilitate furhter research on AI solutions intentionally designed to address the diverse needs and educational contexts of students in LMICs.

The subsequent empirical work of evaluating the ability of generative LLMs to grade student responses in the dataset, aims to better understanding whether LLMs can be used to support ASAG. As discussed above, there is limited empirical work evaluating the use of generative LLMs for ASAG, and while the models appear to be extremely capable with classification tasks, they were not explicitly trained for them, so rigorously testing their capabilities is both under-researched and practically important. Additionally, because this dataset has linked demographic information for each response, having various versions of LLMs mark the responses will allow us to investigate whether the model marks students' answers based on different background attributes (age, gender, home language ability) relative to human raters.

If generative LLMs demonstrate high accuracy and reliability, and don't introduce unintentional bias into the marking process, this would have important practical implications. This is primarily because generative LLMs are relatively straightforward to use and do not require the technical expertise and medium-sized datasets necessary for task-specific fine-tuning approaches, such as transfer learning, meaning they could realistically be used at the school and/or classroom level.

### 3.1 New Dataset

Through a collaboration between researchers at the University of Oxford and Rising Academies (an education organization in Ghana that runs and supports schools), a new reading comprehension dataset has been created. The University of Oxford & Rising Academies Ghanaian Student Reading Comprehension Short Answer Dataset (subsequently referred to as the Ghana Dataset) contains the results of a reading comprehension assessment taken by 162 students between the ages of 13 and 18. The passages, story, and questions were all released items from the 2016 PrePIRLS an international comparative assessment that measures the reading ability of fourth graders across 60 countries explicitly designed for use in LMICs and countries with lower reading levels (Methods and Procedures in PIRLS 2016, 2018)

**Table 2**

*Key Characteristics of Ghana Dataset*

| | |
|:---:|:---:|
| **Total Students** | 162 |
| **Pre-readers vs Readers** | 32 / 130 |
| **Students Completing Written Assessment** | 130 |
| **Students Completing Spoken Assessment** | 48 |
| **Age**: Average, min, max | 13, 9, 18 |
| **Speak English at Home**: count (percent) | 16 (12%) |
| **Female vs Male**: count (percent) | 86 (65%) / 46 (35%) |
| **Total Responses** | 1,068 |
| **Written Responses vs Spoken Responses** | 768 / 300 |
| **Correct vs Incorrect Responses** | 703 / 365 |

To evaluate reading comprehension, students were first given an oral reading fluency task, those determined to be non-readers (32) where they did not take longer silent reading exams. The remaining 130 students were then asked to silently read a 400-word fictional story and answer six short-answer questions. All written responses were scanned and transcribed by expert human annotators. Of those 132 students, 48 students were asked to read a similar short story silently and then answered the 10 reading comprehension questions aloud rather than in writing. This was done to further analyze reading ability separate from writing skills. Their spoken responses were recorded and transcribed.

Once all responses were transcribed, two raters marked each of the 1,068 student answers in the dataset. The raters first read the two stories and then had access to the relevant passage, the student answer, and some examples of correct and incorrect answers. They were asked to rate each response on a binary rubric (correct | incorrect) and a three-class rubric (correct | partially correct | incorrect). As the questions were primarily information retrieval and direct comprehension questions, the raters were given a minimal rubric, instructing them to ignore small spelling and grammar issues and determine whether the student's answer demonstrated that they had understood the question and answered it correctly. In the case of the three-class rating, they were told the following: "*An answer is wrong if it definitely does not answer the question correctly. An answer is partially correct if it gets part of the answer but leaves out important information. An answer is correct if it answers the question and includes the key information.*"

These initial ratings were used to calculate inter-rater agreement scores to better understand the extent to which there was inherent ambiguity or difference in opinion about specific answers. To establish ground truth scores against which the models' performance would be evaluated, the lead author also rated the questions. While the overwhelming majority of answers had unanimous agreement (approximately 90%), in cases where the first two raters disagreed, the third rater's score was used to cast a tie-breaking vote, which was used as the ground truth for whether a student answer was correct or incorrect.

| Passage |
|---|
| "The river is flooding," said the giraffe. "A wall of water is racing down the valley and will soon be here." "What can we do?" asked the gazelle. "It's too late to run away." "Climb up here," called the monkey from the treetops. "The river won't reach the high branches." The animals raced to the trees. But some of them could not climb up the slippery tree trunks. Their hooves and tails were not made for climbing. |
| **Question** |
| Why were the animals trying to climb to the treetops? |
| **Student Answer 1** |
| *The reason why they were trying to climb to the treetops is because the river won't reach the high branches* |

| | Rater Assigned Scores | |
|---|---|---|
| | 2 - class | 3 - class |
| **Rater 1** | correct | partially correct |
| **Rater 2** | incorrect | incorrect |
| **Rater 3** | correct | partially correct |
| **Ground Truth** | correct | partially correct |

| **Student Answer 2** |
|---|
| *The flooded river was splashing around the animals.* |

| **Rater Assigned Scores** | | |
|---|---|---|
| | **2 - class** | **3 - class** |
| **Rater 1** | correct | correct |
| **Rater 2** | correct | correct |
| **Rater 3** | correct | correct |
| **Ground Truth** | correct | correct |

**Figure 2** *Examples of Student Answers and Ratings*

## 3.2 Measuring Model Performance

A set of widely used metrics in machine learning research are precision, recall, and F1 score. These measures allow estimates of model accuracy after accounting for imbalanced classes in the dataset (Banerjee et al., 2008). However, these scores also have limitations: first, they only work for accuracy measures rather than agreement measures because the metric requires assuming that one set of ratings is the "ground truth". Secondly, because the metric only evaluates prediction accuracy, it does not account for true negatives, which can result in a misleadingly high score for datasets with a large proportion of true negatives (Belur et al., 2018).

For these reasons, we also report the Kappa scores, which are chance-adjusted metrics of agreement. These metrics report values ranging from -1 to 1, with a value of 1 indicating perfect agreement beyond what would be expected by chance, a value of 0 suggesting that the agreement is only what would be expected by chance, and a value less than 0 indicating agreement worse than random chance. There are several different measures of chance-adjusted agreement, including Cohen's Kappa (for two raters), the Fleiss Kappa (an adaptation of Cohen's Kappa for 3 or more raters), the Pearson r, and Krippendorff's alpha. We choose to report Cohen's Kappa as it is well-suited for situations where all items are rated by two raters or when model predictions are being compared against the ground-truth values (Banerjee et al., 2008). When evaluating 2-class ratings (incorrect/correct), we used Linear Weighted Kappa (LWK), and when evaluating 3-class ratings, we used Quadratic Weighted Kappa (QWK). The use of QWK is recommended when a rating scale is ordinal, and we believe the 3-class incorrect, partially correct, correct condition is best understood as an ordinal scale (De Raadt et al., 2021).

In some cases, it is possible for the model's Kappa score in relation to the ground truth score to exceed interrater agreement between different human judges. This is mathematically possible because the ground truth answer, in some cases, represents an approximation of the consensus view on the dataset. This is similar in many tasks that involve some degree of human judgment. Even in seemingly more objective tasks such as audio transcription, the ground truth transcript is often established by combining raters' judgments or asking a single rater to transcribe a file multiple times with great care. To establish human-level performance relative to the definitive ground truth answer, it is typical to calculate an individual rater's agreement with the ground truth answer. A model can be said to exceed human-level performance when it corresponds more closely to the ground truth answer (generated by a consensus of human raters) than an individual rater corresponds with the ground truth.

### 3.3 Model Selection and Task Design

Hence the experimental part of this paper focuses on five interconnected research questions, including: (1) How well does the best performing LLM mark student short answer responses relative to human raters? (2) What types of straightforward prompting strategies produce the best results? (3) How large is the performance drop between the top-performing model and the previous generation of models? (4) Do the models demonstrate any converging behaviors in relation to intrarater reliability (i.e., does the model grade a given response the same each time)? (5) Does the model demonstrate any differential performance based on the gender, age, or home language of student.

At the time of writing, the strongest performing model publicly available across a variety of benchmarks was GPT-4 -0613. Hence, we decided to use this model as our upper benchmark. To test the performance of a previous generation model and better understand recent performance improvements, we also used GPT-3.5 turbo. We compared the performance of four variations of GPT to estimate both the overall impact of using generative LLMs compared to transfer learning LLMs and to understand the relative impact of different model sizes and prompt engineering strategies. In all cases, we set the temperature of the model to zero, as this decreases the variability of the model outputs and was recommended for classification tasks by the OpenAI team (personal correspondence).

Every item in the Ghana dataset included: (1) a passage, (2) a comprehension question, (3) a student answer, (4) a human-generated ground truth score on a 2-point scale, (5) a human-generated ground truth score on a 3-point scale, (6) the student's age, (7) gender, and (8) whether they spoke English at home. For each task, the models were presented with the passage, question, and candidate answer and tasked with predicting whether the candidate answer was correct or not. These model predictions were then compared to the ground truth labels generated by expert human raters. The model was only ever shown the passage, question, and student answers; it was never shown the ground-truth answer or any student demographic information.

### 3.4 Prompting Strategy

We employed a relatively simple prompting strategy, as we were trying to simulate what might be realistic for a teacher to create to help them mark a couple hundred formative assessment questions. Our prompting strategy included a system prompt that gave explicit instructions about the more general task and a user prompt that provided the specific passage, question, and answer to evaluate. As shown in Figure X below, we didn't construct a detailed rubric; we simply tasked the model with deciding if the student's answer was correct or incorrect. In the case of the three-class evaluation, we provided the model with a slightly more detailed rubric.

| System Prompt - 2 Class | System Prompt - 3 Class |
| --- | --- |
| You are evaluating students' reading comprehension. | You are evaluating students' reading comprehension. |
| Students have been asked to read a short story and answer reading comprehension questions. | Students have been asked to read a short story and answer reading comprehension questions. |
| You will be given a story, a question, and a student response. | You will be given a story, a question, and a student response. |
| Your task is to classify student responses as correct or incorrect. | Your task is to classify student responses as wrong, partially correct, or correct. |
| You will be given a story, a question, and a student response. | An answer is wrong if it does not answer the question correctly. |
| If the student's response is correct you will respond "1". | An answer is partially correct if it gets part of the answer, but leaves out important information |
| If the student's response is wrong you will respond "0". | An answer is correct if it answers the question and includes the key information |
| | If the student's response is wrong you will respond "0". |
| | If the student's response is partially correct you will respond "1" |
| | If the student's response is correct you will respond "2". |

**Figure 3** *System Prompts Used in Experiment*

We also experimented with the amount of the story we presented to the model (the entire story vs. the relevant passage) and whether we gave the model examples of correct and incorrect answers (i.e., few-shot learning) or no examples (zero-shot). For few-shot learning, we provided examples of a passage, a question, and an example answer from each class. To conduct the model evaluation, we created a simple Python script to (a) pull the relevant passage, question, and candidate answer from our dataset (b) insert them at the appropriate places in the prompt, (c) pass each complete prompt to the OpenAI API, and (d) record the response (i.e. the predicted class).

## 4. Results

### 4.1 Model Performance

Tables 3 and 4 below report the model performance across the 8 pertinent combinations. These results are relative to the human-generated ground truth answer, so a score of 1.00 would indicate that the model gave the same answer as the human grader on each of the 1,068 questions. As discussed above in Section 3.3, while precision, recall, and F1 scores offer valuable insights into the model performance, our preferred holistic measure of model performance is the Kappa score. We use Quadratic Weighted Kappa (QWK) for the 3-class score because we consider it an ordinal ranking, and Linear Weighted Kappa (LWK) for the two-class scores. While we experimented with 16 different combinations of model versions, prompting strategies, and answer classes, we found no meaningful difference between including the entire story vs. the specific passage that contained the information relevant to the answer, and have removed it from the below results for clarity.

**Table 3**

*Performance of Four Varieties of GPT Models on 2-Class Task*

|  | **Prediction** | **Precision** | **Recall** | **F1** | **LWK** |
|---|---|---|---|---|---|
| **basic-zeroshot-3.5** | *Incorrect* | 0.87 | 0.81 | 0.84 | |
|  | *Correct* | 0.91 | 0.93 | 0.92 | 0.76 |
|  | *Average* | 0.89 | 0.89 | 0.89 | |
| **basic-fewshot-3.5** | *Incorrect* | 0.84 | 0.85 | 0.84 | |
|  | *Correct* | 0.92 | 0.92 | 0.92 | 0.76 |
|  | *Average* | 0.89 | 0.89 | 0.89 | |
| **basic-zeroshot-4** | *Incorrect* | 0.87 | 0.99 | 0.91 | |
|  | *Correct* | 1.00 | 0.90 | 0.95 | 0.86 |
|  | *Average* | 0.94 | 0.93 | 0.93 | |
| **basic-fewshot-4** | *Incorrect* | 0.87 | 0.99 | 0.93 | |
|  | *Correct* | 1.00 | 0.92 | 0.96 | 0.89 |
|  | *Average* | 0.95 | 0.95 | 0.95 | |

We find that GPT-4 consistently outperforms GPT-3.5 on Kappa and F1 scores in both the 2-class and 3-class conditions. We also found that few-shot prompting improves performance across conditions, which is consistent with prior research on prompting strategies across a variety of domains. The top-performing

18combination was a few-shot GPT-4, achieving a QWK of 0.92 in the 3-class condition and an LWK of 0.89 in the 2-class condition.

**Table 4**

*Performance of Four Varieties of GPT Models on 3-Class Task*

|  | Prediction | Precision | Recall | F1 | QWK |
|---|---|---|---|---|---|
| **basic-zeroshot-3.5** | *Incorrect* | 0.92 | 0.57 | 0.71 | 0.77 |
|  | *Partial* | 0.22 | 0.71 | 0.33 |  |
|  | *Correct* | 0.93 | 0.80 | 0.86 |  |
|  | *Average* | 0.86 | 0.72 | 0.76 |  |
| **basic-fewshot-3.5** | *Incorrect* | 0.85 | 0.82 | 0.83 | 0.78 |
|  | *Partial* | 0.20 | 0.10 | 0.14 |  |
|  | *Correct* | 0.85 | 0.93 | 0.89 |  |
|  | *Average* | 0.79 | 0.82 | 0.80 |  |
| **basic-zeroshot-4** | *Incorrect* | 0.88 | 0.98 | 0.93 | 0.88 |
|  | *Partial* | 0.33 | 0.49 | 0.39 |  |
|  | *Correct* | 0.96 | 0.83 | 0.89 |  |
|  | *Average* | 0.88 | 0.85 | 0.86 |  |
| **basic-fewshot-4** | *Incorrect* | 0.91 | 0.99 | 0.95 | 0.92 |
|  | *Partial* | 0.39 | 0.39 | 0.39 |  |
|  | *Correct* | 0.95 | 0.91 | 0.93 |  |
|  | *Average* | 0.89 | 0.88 | 0.88 |  |

Another noteworthy trend is that the models' performance with partial credit scores was consistently worse than their performance on fully correct or fully incorrect answers, suggesting that the models struggle more with identifying partially correct responses compared to the other two categories.

Finally, the improvement in performance from GPT-3.5 to GPT-4 is particularly striking, with GPT-4 consistently outperforming its predecessor across all metrics and classification tasks. A closer examination of the results reveals that the performance jump from GPT-3.5 to GPT-4 seems to have mainly come from more accurately grading incorrect answers, as evidenced by the much larger increases in F1 scores for the incorrect class between the two model versions. This suggests that GPT-4 has developed a better understanding of what constitutes an incorrect response, leading to improved overall performance.

## 4.2 Interpretation

In terms of interpreting the strength of these results, there are a few potential approaches. The first and perhaps most general approach is to use established ranges considered to indicate a satisfactory level of agreement. While there are several limitations with this approach, Kappa scores above 0.5 are considered meaningful, above 0.6 as substantial, and Krippendorff claims that alpha scores between 0.67 and 0.80 can be used for drawing provisional conclusions (Landis & Koch, 1977). A slightly better approach is to compare model performance against similar models' performance on tasks that are similar in type (i.e., short answer) or domain (i.e., reading ability). This has the benefit of being able to make a direct comparison but has the shortcoming of not accounting for the fact that different tasks may have different levels of complexity, and that QWK, while it attempts to adjust for the number of classes, typically produces higher scores for tasks with fewer classes. One near comparison would be the results on the NAEP reading comprehension challenge discussed above in section 2.3. The QWK score for the winning submission was 0.84. Another potential point of comparison is the degree of inter-rater agreement between expert raters when evaluating student reading fluency for other items on NAEP. Various studies have reported inter-rater reliability ranging between 0.74 and 0.82 on this task (Smith & Paige, 2019). Using either of these approaches, our results of a QWK of 0.92 in the 3-class condition and an LWK of 0.89 in the 2-class condition would be considered strong results. Another way to evaluate the model performance is to compare it to the rate of agreement between human raters. While we discuss this more extensively above in section 3.3, the intuition is that inter-rater agreement effectively sets a ceiling on the accuracy we could expect from a model.

**Table 5**

*Performance Across Different Approaches to Scoring Responses*

|  | **GPT 4 Few-Shot** | **Expert Human Raters** |
|---|---|---|
| ***3 - Class** (QWK)* | 0.92 | 0.91 |
| ***2 - Class** (LWK)* | 0.89 | 0.96 |

The model's performance is strong here as well, with that the model's agreement level with the ground truth answers exceeded the agreement level of expert human raters in the case of a three-class judgment. To our knowledge, this is the first reported result of an Automated Short Answer Grading (ASAG) model matching or exceeding human agreement levels when evaluating open-response reading comprehension questions. In sum, these results can be considered as strong model performance. Furthermore, because the Ghana Dataset was recently compiled by the authors and is not publicly available, the results would not have been part of prior GPT training runs, which reduces the potential of the LLM having "learned" the correct answers.

## 4.3 Evaluating Model Reliability

Whereas feature-based models and fine-tuned LLMs are constructed in such a way that they must produce an output, given an input, Generative LLMs technically generate their outputs as part of a text output, and can produce different outputs given the same inputs. While setting the model temperature to zero, in principle, freezes the model so it will produce the same output, using APIs means we do not have full control of these models for automated grading, and many of the large LLM providers are regularly tinkering with parameters of their models, so there is the possibility that models will change their results from one time to the next. Interestingly, this is quite like human graders, who occasionally will award different ratings to the same student response, when asked to re-rate it after a period of time. Measures of intrarater reliability are intended to evaluate the extent to which a single rater agrees with their own judgment over time.

To investigate the intrarater reliability of these models, we had the best-performing configurations re-mark the entire dataset and compared the results, on the 2-class task. We observed high but not perfect rates of agreement, with seven students' answers having different labels applied to them. While a Cohen's Kappa score of 0.98 would be considered exceptionally high for human raters, the lack of perfect agreement was an intriguing finding, given the model had the exact same inputs, and had its temperature set to zero, which typically is considered to eliminate the variability in the output.

> *Because it has a long neck*
> *They get down to the grass*
> *The giraffe was tall.*
> *they slide on his back and fall on the ground*
> *the rainy season*
> *The shepherd give Lucy a big smile because it ...*

**Figure 4**  *Student Responses That Were Classified Differently*

A qualitative inspection of the student responses which were classified differently, Figure 4, did not reveal any obvious attributes. Our best current explanation for this variability is that OpenAI is understood to be using an approach called "mixture of experts" whereby API calls are first screened and routed by intermediate models, and then sent to a model optimized for the task, which might introduce a small amount of variability into the system.

## 4.4 Evaluating Potential Bias

Bias and fairness in AI models, particularly in the context of education, have been extensively studied in the literature (see Baker for a comprehensive review). While one potential benefit of using models to mark student work is increased objectivity, modern LLMs are explicitly trained to avoid prejudice. However, the definition of fairness and bias in the AI literature is complex and often contested. In this context, we define bias as the model exhibiting judgment that is significantly different from human raters, based on a student's membership in a salient socio-economic group. For example, if the model systematically rated responses from female students lower than those of male students, in comparison to the scores given by human raters, we might suspect the model was exhibiting a form of bias. The underlying mechanisms by which a generative LLM might operate in this way are complex and open for debate, especially because we provide the same prompt for all students and the model has no access to student demographic information. However, it is possible, in principle, that due to the corpus on which the model was trained, it might "prefer" certain types of student responses, such as those of native English speakers, over those of non-native speakers. If this preference differed from that of expert human raters, it would be a cause for concern. It is also possible that such a discrepancy could point to implicit biases or prejudices held by the human raters themselves, although a proper treatment of this question is outside the scope of this article.

To investigate potential biases in the model, we defined model misclassification as cases where the model's prediction did not match the ratings provided by human raters. We then examined the misclassification rates by gender and whether a student had some exposure to the English language at home. More formally, we compared the probability of the model misclassifying a student's answer (i.e., not matching the human-generated ground-truth rating) conditional on the student being male to the probability of misclassification for female students. We made the same comparison between students who spoke English at home and those who did not.

**Table 5**

*Misclassification Rates by Demographic Group*

|  | Whole Dataset | Misclassified Responses |
|---|---|---|
| **Male** | 33% \| (354) | 35% \| (9) |
| **Female** | 67% \| (714) | 65 % \| (17) |
| **English at Home** | 13% \| (144) | 8 % \| (2) |
| **No English at Home** | 87% \| (924) | 92 % \| (24) |

We found no statistically significant difference due to socioeconomic factors. While this is encouraging, given the relatively small numbers of total misclassifications (26), due the practical and ethical implications of this question, we treat our results with caution, and believe that these types of checks should be conducted regularly.

## 5. Discussion

Our topline finding, that GPT-4 can reach equivalent performance levels to expert human raters on ASAG tasks, is a substantial advance, and has important implications for use in real-world contexts. Human ratings are currently the gold standard for assessing model performance on short answer questions, and human raters are also the norm in real-world educational settings (i.e., schools, tutoring sessions). If LLMs can perform as well as human raters, as our research suggests, it indicates that they could form the basis of assessment systems that could be used in schools.

Importantly, these findings were achieved with no fine-tuning and limited prompt engineering, suggesting that generative LLMs could be useful for low-stakes assessment tasks in real-world educational settings. This could have significant benefits, particularly in places with limited assessment practices or where fine-tuning is not feasible, such as resource-constrained environments. The top performance level was achieved with minimal additional model configuration (or fine-tuning), which points to their potential for use in settings where there is limited technical expertise. If the assessment of open-ended responses could be automated, even partially - which our research suggests could be possible - the reliable assessment of responses to short answer questions becomes feasible at a large scale.

### 5.1 Implications

Implementing an assessment system that leverages LLMs for formative assessment purposes in real-world situations would require some technical infrastructure. In high-income countries, we believe there is significant potential in the short term for using LLMs to support more frequent and effective formative assessments. This could potentially have a disproportionate benefit for lower-resourced schools, where the quantity and quality of formative assessment is typically lower. While our findings suggest that these approaches would work with student work from anglophone LMICs, implementing them in practice would still face challenges. It is unlikely that individual schools in LMICs could execute this without support, particularly for formative assessment. For example, schools would need to be provided with the appropriate devices, infrastructure, and connectivity, and teachers and school staff would need to be trained. In the near term, a more likely use case in LMIC contexts would be for large-scale assessments meant to screen for student ability and mastery of basic skills.

The current cost of accessing GPT-4, as well as the technical expertise required to interact with the APIs of these LLMs, should be taken into account when considering how to take this work forward. While both the costs of using these models and the tools being developed for non-expert users are advancing rapidly, at the time of writing, there would still be significant barriers to generative LLMs being used at scale to support the marking of formative assessments, particularly in low-resource settings. The financial and technical constraints could hinder the widespread adoption of these technologies, especially in educational contexts where resources are already limited. Better understanding these practical limitations will be crucial for realizing the potential benefits of generative LLMs in automated short answer grading and ensuring that these advancements are accessible to a wide range of educational institutions and learners.

## 5.2 Limitations

While this study provides valuable insights into the potential of generative LLMs for automated short answer grading, it is important to acknowledge its limitations. While the Ghana Dataset fills an important gap as it allows for the evaluation of an underrepresented group, it ultimately consists of student responses in English from a single country. It is unclear how well the model's performance would generalize to other languages, dialects, or cultural contexts. Future research should explore the performance of generative LLMs across a wider range of languages and cultural settings to better understand their potential for global application in education; it may also limit the generalizability of the findings to other educational contexts and student populations. Additionally, while the Ghana Dataset provides some demographic information about the students, such as age, gender, and whether they speak English at home, it may not capture the full range of socioeconomic and cultural factors that could influence student performance.

In terms of the experimental section, a key limitation is the type and difficulty of the reading comprehension questions used. The questions were primarily direct comprehension questions, assessing students' ability to retrieve explicit information from the given passages. It remains uncertain how the model's performance would change as the complexity of the topics and the difficulty of the questions increase. Furthermore, the study did not explore the model's performance across different subject areas, which could have varying levels of complexity and require different types of reasoning skills. Additionally, the study employed simple binary and three-class rubrics that primarily focused on overall correctness. It is unclear how the model would perform if the scoring criteria were more complex or had more potential classes, such as those that assess the quality of the student's writing or the depth of their understanding. Finally, it is worth noting that this study did not evaluate multiple generative LLMs, focusing solely on GPT-4. Given the rapid pace of development in the field of AI, it is likely that the performance of these models will continue to improve over time. As such, the results of this study should be interpreted as a snapshot of the current state of the technology rather than a definitive assessment of its potential.

## 5.3 Further Research

Further research should focus on expanding the scope and diversity of datasets used to evaluate the performance of generative LLMs in automated short answer grading. This includes exploring datasets that cover a wider range of grade levels, subject areas, and question types, as well as varying levels of difficulty. By testing the models on a more diverse set of educational contexts, researchers can better understand the generalizability and limitations of these technologies. Additionally, future studies should investigate the performance of generative LLMs on datasets from non-English languages, as the current study focused solely on English-language responses from Ghanaian students. Evaluating the models' performance across different languages and cultural contexts is crucial for determining their potential for global application in education.

Another important area for further research is the ethical implications of using generative LLMs for automated grading. While the current study briefly addresses the topic of bias in the model's performance, it does not fully explore the potential concerns regarding privacy, fairness, and the perpetuation or amplification of existing educational inequities. Future research should delve deeper into these ethical considerations and develop guidelines for the responsible use of generative LLMs in educational assessment. Moreover, further investigation into the intrarater reliability of model results is necessary, as ensuring the consistency and trustworthiness of the grading outcomes is critical for educators to have confidence in the technology. Finally, while this study focused on GPT-4, it is essential to evaluate the performance of other generative LLMs, including open-source models that could potentially be more affordable and transparent. By exploring a wider range of models and their capabilities, researchers can better inform the development and implementation of these technologies in real-world educational settings.

## 6 Conclusion

This study aimed to address the lack of empirical research on the use of generative Large Language Models (LLMs) for educational assessment, particularly in the context of low and middle-income countries (LMICs). To achieve this goal, we introduced a novel dataset, the University of Oxford & Rising Academies Ghanaian Student Reading Comprehension Short Answer Dataset (Ghana Dataset), containing over 1,500 student answers to reading comprehension questions from 130 students aged 9 to 18 in Ghana. This dataset provides a unique opportunity to evaluate the performance and potential of generative LLMs in a diverse educational context, while also offering insights into the educational challenges faced by students in LMICs.

Our experimental design focused on evaluating the performance of GPT-4, a state-of-the-art generative LLM, in grading short answer reading comprehension questions from the Ghana Dataset. We employed various prompting strategies, including zero-shot and few-shot learning, and compared the model's performance to that of expert human raters. The results showed that GPT-4, with minimal prompt engineering, achieved near-parity with expert human raters, attaining a Quadratic Weighted Kappa (QWK) score of 0.92 in the 3-class condition and a Linear Weighted Kappa (LWK) score of 0.89 in the 2-class condition. These findings demonstrate the remarkable ability of generative LLMs to evaluate student responses accurately, even when dealing with a novel dataset from an underrepresented educational context. The implications of these findings are significant for both high-income countries and LMICs. In high-income countries, generative LLMs could be used to support more frequent and effective formative assessments, particularly in lower-resourced schools where the quantity and quality of formative assessment are typically lower. In LMICs, while implementing these approaches may face challenges at the individual school level, they could be valuable for large-scale assessments aimed at screening student ability and mastery of basic skills. However, it is important to note that implementing these approaches in real-world situations would require some technical infrastructure and support.

This study contributes to the growing body of research on the application of generative LLMs in education by providing empirical evidence of their potential for automated short answer grading. The introduction of the Ghana Dataset also helps to address the scarcity of diverse and comprehensive datasets of student short answer questions, particularly from LMICs. As the field of AI in education continues to evolve, it is crucial to ensure that the benefits of these technologies are accessible to students from all backgrounds and educational contexts. Future research should further explore the potential of generative LLMs in various educational settings, while also addressing the challenges and ethical considerations associated with their implementation.

**Declarations**

One of the authors has an ongoing research partnership with Rising Academies.

# References

Alikaniotis, D., Yannakoudakis, H., & Rei, M. (2016). Automatic Text Scoring Using Neural Networks. *Proceedings of the 54th Annual Meeting of the Association for Computational Linguistics (Volume 1: Long Papers)*, 715–725. https://doi.org/10.18653/v1/P16-1068

Baker, R. S., & Hawn, A. (2022). Algorithmic Bias in Education. *International Journal of Artificial Intelligence in Education*, *32*(4), Article 4. https://doi.org/10.1007/s40593-021-00285-9

Banerjee, M., Capozzoli, M., Mcsweeney, L., & Sinha, D. (2008). Beyond Kappa: A Review of Interrater Agreement Measures. *Canadian Journal of Statistics*, *27*, 3–23. https://doi.org/10.2307/3315487

Bellinger, J. M., & DiPerna, J. C. (2011). Is fluency-based story retell a good indicator of reading comprehension? *Psychology in the Schools*, *48*(4), Article 4. https://doi.org/10.1002/pits.20563

Belur, J., Tompson, L., Thornton, A., & Simon, M. (2018). Interrater Reliability in Systematic Review Methodology: Exploring Variation in Coder Decision-Making. *Sociological Methods & Research*, *50*, 004912411879937. https://doi.org/10.1177/0049124118799372

Black, P., & Wiliam, D. (2009). Developing the theory of formative assessment. *Educational Assessment, Evaluation and Accountability*, *21*(1), Article 1. https://doi.org/10.1007/s11092-008-9068-5

Bommasani, R., Hudson, D. A., Adeli, E., Altman, R., Arora, S., von Arx, S., Bernstein, M. S., Bohg, J., Bosselut, A., Brunskill, E., Brynjolfsson, E., Buch, S., Card, D., Castellon, R., Chatterji, N., Chen, A., Creel, K., Davis, J. Q., Demszky, D., … Liang, P. (2022). *On the Opportunities and Risks of Foundation Models* (arXiv:2108.07258; Issue arXiv:2108.07258). arXiv. https://doi.org/10.48550/arXiv.2108.07258

Brown, T. B., Mann, B., Ryder, N., Subbiah, M., Kaplan, J., Dhariwal, P., Neelakantan, A., Shyam, P., Sastry, G., Askell, A., Agarwal, S., Herbert-Voss, A., Krueger, G., Henighan, T., Child, R., Ramesh, A., Ziegler, D. M., Wu, J., Winter, C., … Amodei, D. (2020). Language Models are Few-Shot Learners. *arXiv:2005.14165 [Cs]*. http://arxiv.org/abs/2005.14165

Burrows, S., Gurevych, I., & Stein, B. (2015). The Eras and Trends of Automatic Short Answer Grading. *International Journal of Artificial Intelligence in Education*, *25*(1), Article 1. https://doi.org/10.1007/s40593-014-0026-8

Caines, A., Benedetto, L., Taslimipoor, S., Davis, C., Gao, Y., Andersen, O., Yuan, Z., Elliott, M., Moore, R., Bryant, C., Rei, M., Yannakoudakis, H., Mullooly, A., Nicholls, D., & Buttery, P. (2023). *On the application of Large Language Models for language teaching and assessment technology* (arXiv:2307.08393). arXiv. http://arxiv.org/abs/2307.08393

Camus, L., & Filighera, A. (2020). Investigating Transformers for Automatic Short Answer Grading. In I. I. Bittencourt, M. Cukurova, K. Muldner, R. Luckin, & E. Millán (Eds.), *Artificial Intelligence in Education* (Vol. 12164, pp. 43–48). Springer International Publishing. https://doi.org/10.1007/978-3-030-52240-7_8

Chowdhery, A., Narang, S., Devlin, J., Bosma, M., Mishra, G., Roberts, A., Barham, P., Chung, H. W., Sutton, C., Gehrmann, S., Schuh, P., Shi, K., Tsvyashchenko, S., Maynez, J., Rao, A., Barnes, P., Tay, Y., Shazeer, N., Prabhakaran, V., … Salakhutdinov, R. (2023). *PaLM: Scaling Language Modeling with Pathways*.

Cohn, C., Hutchins, N., Le, T., & Biswas, G. (2024). *A Chain-of-Thought Prompting Approach with LLMs for Evaluating Students' Formative Assessment Responses in Science* (arXiv:2403.14565). arXiv. http://arxiv.org/abs/2403.14565

De Raadt, A., Warrens, M. J., Bosker, R. J., & Kiers, H. A. L. (2021). A Comparison of Reliability Coefficients for Ordinal Rating Scales. *Journal of Classification*, *38*(3), Article 3. https://doi.org/10.1007/s00357-021-09386-5

Fernandez, N., Ghosh, A., Liu, N., Wang, Z., Choffin, B., Baraniuk, R., & Lan, A. (2023). *Automated Scoring for Reading Comprehension via In-context BERT Tuning* (arXiv:2205.09864; Issue arXiv:2205.09864). arXiv. http://arxiv.org/abs/2205.09864

Gilardi, F., Alizadeh, M., & Kubli, M. (2023). ChatGPT outperforms crowd workers for text-annotation tasks. *Proceedings of the National Academy of Sciences*, *120*(30), Article 30. https://doi.org/10.1073/pnas.2305016120

Haller, S., Aldea, A., Seifert, C., & Strisciuglio, N. (2022). *Survey on Automated Short Answer Grading with Deep Learning: From Word Embeddings to Transformers* (arXiv:2204.03503). arXiv. http://arxiv.org/abs/2204.03503

Kojima, T., Gu, S. S., Reid, M., Matsuo, Y., & Iwasawa, Y. (2022). *Large Language Models are Zero-Shot Reasoners*.

Kortemeyer, G. (2023). *Performance of the Pre-Trained Large Language Model GPT-4 on Automated Short Answer Grading* (arXiv:2309.09338). arXiv. http://arxiv.org/abs/2309.09338

Kuzman, T., Mozetič, I., & Ljubešić, N. (2023). *ChatGPT: Beginning of an End of Manual Linguistic Data Annotation? Use Case of Automatic Genre Identification* (arXiv:2303.03953; Issue arXiv:2303.03953). arXiv. http://arxiv.org/abs/2303.03953

Landauer, T. K., Lochbaum, K. E., & Dooley, S. (2009). A New Formative Assessment Technology for Reading and Writing. *Theory Into Practice*, *48*(1), Article 1. https://doi.org/10.1080/00405840802577593

Landis, J. R., & Koch, G. G. (1977). The Measurement of Observer Agreement for Categorical Data. *Biometrics*, *33*(1), Article 1. https://doi.org/10.2307/2529310

Magliano, J. P., & Graesser, A. C. (2012). Computer-based assessment of student-constructed responses. *Behavior Research Methods*, *44*(3), Article 3. https://doi.org/10.3758/s13428-012-0211-3

Matelsky, J. K., Parodi, F., Liu, T., Lange, R. D., & Kording, K. P. (2023). *A large language model-assisted education tool to provide feedback on open-ended responses* (arXiv:2308.02439). arXiv. http://arxiv.org/abs/2308.02439

Mishra, S., Khashabi, D., Baral, C., & Hajishirzi, H. (2022). *Cross-Task Generalization via Natural Language Crowdsourcing Instructions* (arXiv:2104.08773). arXiv. http://arxiv.org/abs/2104.08773

Mizumoto, A., & Eguchi, M. (2023). Exploring the potential of using an AI language model for automated essay scoring. *Research Methods in Applied Linguistics*, *2*(2), 100050. https://doi.org/10.1016/j.rmal.2023.100050

Morjaria, L., Burns, L., Bracken, K., Levinson, A. J., Ngo, Q. N., Lee, M., & Sibbald, M. (2024). Examining the Efficacy of ChatGPT in Marking Short-Answer Assessments in an Undergraduate Medical Program. *International Medical Education*, *3*(1), 32–43. https://doi.org/10.3390/ime3010004

Ouyang, L., Wu, J., Jiang, X., Almeida, D., Wainwright, C. L., Mishkin, P., Zhang, C., Agarwal, S., Slama, K., Ray, A., Schulman, J., Hilton, J., Kelton, F., Miller, L., Simens, M., Askell, A., Welinder, P., Christiano, P., Leike, J., & Lowe, R. (2022). *Training language models to follow instructions with human feedback* (arXiv:2203.02155; Issue arXiv:2203.02155). arXiv. https://doi.org/10.48550/arXiv.2203.02155

Perez, E., Kiela, D., & Cho, K. (2021). *True Few-Shot Learning with Language Models*.

Raffel, C., Shazeer, N., Roberts, A., Lee, K., Narang, S., Matena, M., Zhou, Y., Li, W., & Liu, P. J. (2020). Exploring the Limits of Transfer Learning with a Unified Text-to-Text Transformer. *arXiv:1910.10683 [Cs, Stat]*. http://arxiv.org/abs/1910.10683

Ridley, R., He, L., Dai, X., Huang, S., & Chen, J. (2020). *Prompt Agnostic Essay Scorer: A Domain Generalization Approach to Cross-prompt Automated Essay Scoring* (arXiv:2008.01441). arXiv. http://arxiv.org/abs/2008.01441

Schneider, J., Schenk, B., Niklaus, C., & Vlachos, M. (2023). *Towards LLM-based Autograding for Short Textual Answers*.

Shapiro, E. S., Fritschmann, N. S., Thomas, L. B., Hughes, C. L., & McDougal, J. (2014). Concurrent and Predictive Validity of Reading Retell as a Brief Measure of Reading Comprehension for Narrative Text. *Reading Psychology*, *35*(7), Article 7. https://doi.org/10.1080/02702711.2013.790328

Shute, V. J. (2008). Focus on Formative Feedback. *Review of Educational Research*, *78*(1), 153–189. https://doi.org/10.3102/0034654307313795

Smith, G., & Paige, D. (2019). A STUDY OF INTERRATER RELIABILITY 1 A Study of Reliability Across Multiple Raters When Using the NAEP and MDFS Rubrics to Measure Oral Reading Fluency. *Reading Psychology*, *40*. https://doi.org/10.1080/02702711.2018.1555361

Stiennon, N., Ouyang, L., Wu, J., Ziegler, D. M., Lowe, R., Voss, C., Radford, A., Amodei, D., & Christiano, P. (2022). *Learning to summarize from human feedback* (arXiv:2009.01325; Issue arXiv:2009.01325). arXiv. https://doi.org/10.48550/arXiv.2009.01325

Sultan, M. A., Sil, A., & Florian, R. (2022). Not to Overfit or Underfit the Source Domains? An Empirical Study of Domain Generalization in Question Answering. *Proceedings of the 2022 Conference on Empirical Methods in Natural Language Processing*, 3752–3761. https://doi.org/10.18653/v1/2022.emnlp-main.247

Sung, C., Dhamecha, T., Saha, S., Ma, T., Reddy, V., & Arora, R. (2019). Pre-Training BERT on Domain Resources for Short Answer Grading. *Proceedings of the 2019 Conference on Empirical Methods in Natural Language Processing and the 9th International Joint Conference on Natural Language Processing (EMNLP-IJCNLP)*, 6070–6074. https://doi.org/10.18653/v1/D19-1628

Wei, J., Tay, Y., Bommasani, R., Raffel, C., Zoph, B., Borgeaud, S., Yogatama, D., Bosma, M., Zhou, D., Metzler, D., Chi, E. H., Hashimoto, T., Vinyals, O., Liang, P., Dean, J., & Fedus, W. (2022). *Emergent Abilities of Large Language Models* (arXiv:2206.07682; Issue arXiv:2206.07682). arXiv. http://arxiv.org/abs/2206.07682

Weidinger, L., Uesato, J., Rauh, M., Griffin, C., Huang, P.-S., Mellor, J., Glaese, A., Cheng, M., Balle, B., Kasirzadeh, A., Biles, C., Brown, S., Kenton, Z., Hawkins, W., Stepleton, T., Birhane, A., Hendricks, L. A., Rimell, L., Isaac, W., … Gabriel, I. (2022). Taxonomy of Risks posed by Language Models. *Proceedings of the 2022 ACM Conference on Fairness, Accountability, and Transparency*, 214–229. https://doi.org/10.1145/3531146.3533088

Ye, Q., Lin, B. Y., & Ren, X. (2021). *CrossFit: A Few-shot Learning Challenge for Cross-task Generalization in NLP* (arXiv:2104.08835). arXiv. http://arxiv.org/abs/2104.08835

Zhao, S., Li, B., Reed, C., Xu, P., & Keutzer, K. (2020). *Multi-source Domain Adaptation in the Deep Learning Era: A Systematic Survey* (arXiv:2002.12169; Issue arXiv:2002.12169). arXiv. http://arxiv.org/abs/2002.12169